# Toward the Axiomatization of Intelligence:
# Structure, Time, and Existence

Kei Itoh*

April 20, 2025

## Abstract

This study aims to construct an axiomatic definition of intelligence within a meta-framework that defines the definition method, addressing intelligence as an inherently naïve and polysemous concept. Initially, we formalize a set-theoretic representation of the universe as the domain wherein intelligence exists and characterize intelligence as a structure that involves temporal evolution and interaction with other sets. Starting from a naïve definition of intelligence as "an entity possessing structures for externally inputting, internally processing, and externally outputting information or matter," we axiomatically reformulate it within this set-theoretical depiction of the universe. Applying this axiomatic definition, we compare and interpret three examples—Hebbian non-optimized neural networks (NNs), backpropagation-optimized NNs, and biological reflexive systems (e.g., the gill-withdrawal reflex of Aplysia)—in terms of their intelligence, structural properties, and biological plausibility. Furthermore, by extending our definition into a categorical framework, we introduce two categories, "Time Category" and "Intelligence Category," along with the functorial relationships between them, demonstrating the potential to represent changes and mimicry relationships among intelligent systems abstractly. Additionally, since intelligence, as defined herein, functions effectively only when accompanied by temporal interactions, we introduce the concept of "activity" and explore how activity-based conditions influence classifications and interpretations of intelligence. Finally, we suggest that our definitional methodology is not limited to intelligence alone but can be similarly applied to other concepts such as consciousness and emotion, advocating their formal reinterpretation through the same procedural steps—definition of a universal representation, selection of naïve definitions, and axiomatic formalization. Consequently, this research serves as a foundational theory for defining intelligence and provides significant implications and a general methodological framework for formally describing various traditional metaphysical concepts.

* Ehime University Graduate School of Science and Technology. The author conducted this research independently while affiliated with Ehime University and utilized the university's library and electronic resources. The research was conducted without specific financial support from external funding sources. The author declares no conflict of interest. While this English version constitutes the formal and citable edition of the work, the original manuscript was composed in Japanese, and some structural or rhetorical nuances may be more fully preserved in the original text.

**Naïve Definitions of Intelligence and Their Examples**

We humans possess "intelligence." Intelligence, or concepts analogous to it, seems to be present not only in humans but also in other living organisms, artificial intelligence composed of computer programs, and even simpler tools—or at least it appears to be so. Throughout history, from ancient times to the present day, numerous people have addressed the question of how to define this concept of "intelligence" or related concepts.

For instance, Plato, in Book VI of "*The Republic*" (e.g., 509d–511e) [1], described true intelligence (*νοῦς*) as the capacity enabling unconditioned contemplation (*νόησις*) of truth and forms, distinct from the capability for mathematical reasoning based on assumptions (*διάνοια*). Kant defined "*Verstand*" (understanding) as "*Verstand ist aber das Vermögen der Begriffe.*" (A69/B94) [2], meaning "Understanding is the faculty of concepts," positioning it as an intellectual function responsible for judgment and synthesis based upon given intuitions. In this regard, "*Verstand*" can be considered a fundamental component of modern intelligence and, thus, constitutes a defining concept of intelligence. From a psychological perspective, Wechsler defined intelligence as "the aggregate or global capacity of the individual to act purposefully, to think rationally, and to deal effectively with his environment" [3]. Furthermore, from a computational and logical standpoint, Turing devised the Turing test to determine "whether something possesses intelligence" [4]. Although this approach intentionally avoids a direct definition, it indirectly defines intelligence, nonetheless. Schrödinger, meanwhile, commented less on intelligence itself and more on the nature of life, stating, "It is by avoiding the rapid decay into the inert state of 'equilibrium' that an organism appears so enigmatic... What an organism feeds upon is negative entropy." [5]

Aside from these examples, numerous definitions of intelligence exist, which indicates that intelligence is typically defined through naïve (non-axiomatic) definitions and is inherently ambiguous. The lack of a unique definition implies the absence of a unified conceptual understanding of "intelligence" among people and societies. Even the five definitions exemplified here—by Plato, Kant, Wechsler, Turing, and Schrödinger—partially correspond at best and cannot be considered fully congruent. The conception of "intelligence," as understood by these individuals, differs significantly and undoubtedly varies among all people. The diversity of intelligence definitions is deeply recognized within the field of artificial intelligence theory; for instance, Legg & Hutter (2007) [6] identified over seventy definitions in the literature, systematically compiling representative definitions. Thus, the question "What is intelligence?" remains an unresolved and fundamental issue, not only within philosophy and psychology but also within the theoretical frameworks of artificial intelligence.

**The Importance of Axiomatic Definition and Mathematical Formalization of Intelligence and Its Capabilities in AI Development**

On the other hand, creating intelligence, including artificial intelligence, has been continuously pursued throughout human (or even biological) history. According to the Law of Accelerating Returns [7], intelligence created by humanity (or life in general) progresses exponentially. This also implies that the intelligence currently being developed by humans, such as Artificial General Intelligence (AGI), exhibits an exponential increase in its level, complexity, and developmental difficulty. Given that intelligence is defined naïvely and ambiguously, significant and various problems can arise in creating such advanced intelligence. Two illustrative examples of this issue are provided below.

In the field of AGI development, there exists a notable conflict between two perspectives: one argues that "extending large language models (LLMs) will achieve high-level intelligence" [e.g., 8,9], while the other maintains that "intelligence cannot be recognized without emphasizing interactions (actions and trial-and-error) between agents and their environments" [e.g., 10,11]. The former prioritizes linguistic processing and logical reasoning capabilities, whereas the latter regards behavioral adaptation as the essence of intelligence. However, both perspectives capture certain aspects of intelligence; their fundamental divergence results from the lack of consensus on the definition of intelligence. Consequently, approaches and evaluation criteria in research differ substantially, hindering mutual understanding and integration. This situation reduces efficiency in developing advanced intelligence and may result in a lack of unified foundational concepts necessary for the societal implementation of artificial intelligence.

Itoh (2024a, 2024b) [12,13] argued that, for artificial neural networks (NNs)-based mimicry of biological neural systems, it is inappropriate to employ biologically implausible structures such as error-optimization methods, including backpropagation. Furthermore, it was demonstrated that directly computing vector similarities using biologically implausible distributed representations is unsuitable for implementing Hebbian learning rules [16] into NNs. Itoh succeeded in developing a more biologically plausible structure, enabling their implementation [13]. This indicates that certain "constraints" exist when creating intelligence for a specific purpose (such as biologically inspired neural AI). However, suppose the definition of intelligence is naïve and ambiguous. In that case, such constraints also become naïve and ambiguous, preventing creators or societies from establishing clear and rigorous definitions of intelligence and its constraints. Consequently, creating intelligence that requires relatively advanced definitions and consensus across society becomes impossible. Furthermore, the prevalence of secondary concepts such as optimization processes [e.g., 17] and backpropagation [e.g., 18]—rather than primary biological concepts like Hebbian learning—in current NNs-AI development theory likely stems mainly from the naïve understanding and ambiguity of intelligence definitions and the lack of axiomatic understanding of its constraints. Thus, developing advanced intelligence and its societal implementation fundamentally requires a sophisticated understanding of intelligence and related constraints. Therefore, moving beyond naïve and ambiguous understandings of intelligence is essential.

**How to Define the "Definition of Intelligence"**

However, it is evident that "intelligence" is inherently a naïve and polysemous concept. Intelligence is neither provable mathematically nor observable physically; rather, it is something humans intuitively understand and individually define. This implies that the "definition of intelligence" is inherently ambiguous. Each individual and society may possess distinct definitions of intelligence; consequently, their methodologies will also differ. Even if someone's definition of intelligence remains undescribed or uncommunicated to others, if it appears self-evident to that individual, then that "definition of intelligence" is effectively correct (or at least irrefutable) from their perspective. Conversely, I argue that to transcend naïve and ambiguous understandings of intelligence, we must establish a meta-definition—essentially, a "definition of how to define intelligence." In other words, we must start by defining the method used to define intelligence. If this "definition of the definition of intelligence" achieves sufficient rigor to enable axiomatic, unambiguous definitions that move beyond naïve interpretations, creating correspondingly advanced forms of intelligence would be possible. Moreover, if such a meta-definition and the resultant definitions of intelligence are logically and scientifically robust, some polysemous interpretations might be acceptable in intelligence creation. Even if multiple definitions of intelligence exist, provided their axiomatic rigor is sufficiently high, they may be fully comparable and corresponded. Additionally, interpreting a specific intelligence and its capabilities through several axiomatic definitions of intelligence could yield significant insights. For example, although distinct, the later-discussed set-theoretic and categorical definitions of intelligence can maintain sufficient mathematical rigor to correspond. In this research, the set-theoretic view focuses on the movement of elements between intelligence and its external environment, whereas the categorical view emphasizes functional correspondence among intelligence. Both perspectives may become essential in various scenarios of intelligence creation.

**Set-Theoretic Depiction of the Universe**

Everything we commonly regard as possessing intelligence exists materially within the universe and has some structure. Naturally, the difference between intelligent and non-intelligent matter should depend on whether their structure qualifies as intelligent. Thus, in formulating a non-naïve definition of intelligence, or rather a definition of how to define intelligence, it becomes necessary to axiomatically define the depiction of the universe in which intelligence exists. Intelligence is defined as specific matter and its structure within a certain universe; hence, defining intelligence fundamentally requires defining the universe itself. In this study, prior to axiomatic definitions of intelligence, we axiomatically define the universal depiction wherein intelligence exists and use this as a basis for defining intelligence. We adopt a "set-theoretic universal depiction," where the universe is axiomatically defined as a set of particles. Within this depiction, "time" is axiomatically defined as a mapping projecting the universal particle set from one point in time to the next. These set-theoretic universal depictions and temporal mappings enable us to define and interpret intelligence and its capabilities. Defining universal depiction and employing this methodology can be regarded as the "definition of how to define intelligence." Here, the rigor of the universal set and temporal mappings must only be sufficient to define the intelligence in question adequately. Thus, certain intelligences may be definable within classical particles and classical mechanics, while others may require expansion into quantum particles or quantum mechanics; the level of detail need only match the required precision. Furthermore, descriptions of particle sets, and dynamic actions can often remain significantly limited. For instance, neural networks (NNs) defined as "intelligent" via programmatic functions or algorithms might adequately be described merely by their mechanisms and operations when considering only their internal existence and capabilities.

**Research Objectives**

Based on the above considerations, this research aims to establish an axiomatic definition of intelligence and provide straightforward interpretations of intelligence capabilities grounded in that definition. The axiomatic definition of intelligence will follow these steps:

1. Axiomatic definition and selection of the set-theoretic universal depiction
2. Selection of a naïve definition of intelligence
3. Axiomatic formalization of the selected naïve intelligence definition based on the set-theoretic universal depiction

Step 2 is necessary due to the inherently naïve nature of "intelligence." Axiomatic intelligence is not defined directly through axioms but by axiomatizing a chosen naïve definition of intelligence. Step 3 demonstrates interpretations of intelligence capabilities through concrete intelligence defined by axiomatic definitions. Specifically, this research applies the axiomatic intelligence definition to Hebbian non-optimized NNs proposed by Itoh (2024b) [13], conventional backpropagation-based optimized NNs [e.g., 19], and the gill-withdrawal mechanism of Aplysia [e.g., 20]. Comparing these examples will elucidate their respective intelligence properties and differences. Finally, we briefly expand the discussion from set theory to category theory, reaffirming the arbitrariness of intelligence definitions, introducing the "activity" derived from intelligence definitions, and contemplating the formalization of other concepts such as consciousness and emotion.

The primary objective of this research is to enable axiomatic definitions of intelligence and its capabilities. However, defining "intelligence" itself holds limited direct practical significance. A practically meaningful definition would pertain to specific intelligence, such as "intelligence X."

Ultimately, we seek to create certain concepts or mechanisms known as intelligence, necessitating their axiomatic definition. In this sense, defining "intelligence" is merely an illustrative example of the broadest possible framework for concepts and structures we aim to create.

**Axiomatic Definition and Selection of the Set-Theoretic Universal Depiction**

In this section, we formalize the universe in which intelligence exists and its temporal evolution using a set-theoretic framework. First, we define the universal set $U$ as the complete set of elements $v$:

$$U := \{v_1, v_2, \cdots, v_{n-1}, v_n\} = \{v | \text{Universe}\}$$

In other words, $U$ is a finite set containing all elements $v$ of the universe. The elements $v$ might be atoms, elementary particles, or even neurons, but detailed descriptions are not necessary here and should instead be specified in concrete definitions of intelligence. Each element $v$ possesses certain states (such as positions or physical quantities), and their detailed descriptions may sometimes be necessary, but again, such specifics should be provided when defining concrete intelligence. The aim of this section is merely to establish the framework of the universe in which intelligence exists. Moreover, because specific descriptions of the universe should differ depending on the intelligence being defined, details concerning elements and their states should accompany those individual intelligence definitions.

The universal set $U$ is projected from one point in time to the next through a temporal mapping $T$. Specifically, the set $U_i$, representing the universe at a particular time $t_i$, evolves into the set $U_{i+1}$ at the next time $t_{i+1}$ through the mapping $T_i$, expressed set-theoretically as follows:

$$T_i: U_i \to U_{i+1}$$

The cardinality of $U$ remains constant. Thus, the passage of time is expressed as a sequence of continuous projections by $T$:

$$T_i: U_i \to U_{i+1},$$
$$T_{i+1}: U_{i+1} \to U_{i+2},$$
$$T_{i+2}: U_{i+2} \to U_{i+3},$$
$$\cdots,$$

Similar to the description of $U$, detailed descriptions regarding $T$—such as considerations of temporal continuity, granularity, or dynamical interactions—are not defined here and should instead be addressed when defining specific intelligences. That is, considerations such as temporal granularity and the specific dynamical actions accompanying time evolution should be tailored individually to each intelligence definition rather than defined in this section.

Additionally, we define and formalize several concepts within this universe for the sake of intelligence definition. Specifically, we set-theoretically formalize the notions of "existence," "structure," and interactions between an existence and its exterior. An existence $E$ within the

universal set $U$ is defined as an arbitrary subset of $U$:
$$E \subseteq U$$
For example, if $E = U$, then $E$ represents the existence of the universe itself. Alternatively, defining the existence of Earth as $E_{\text{Earth}}$, we formalize it as:
$$E_{\text{Earth}} \coloneqq \{\alpha_1, \alpha_2, \cdots, \alpha_{e-1}, \alpha_e\} = \{\alpha | \text{Earth}\} (\text{Earth} \in \text{Universe})$$
Clearly, Earth is not identical to the entire universe, and thus $E_{\text{Earth}} \subset U$. The method for specifying all elements belonging to Earth (or any other existence) should be detailed similarly to elements and time, accompanying specific definitions of intelligence.

An existence has a certain structure $C$, defined by relationships (such as relative positions) among its elements, formalized using Cartesian product sets as follows. For an existence $E = \{\varepsilon_1, \varepsilon_2, \cdots, \varepsilon_{j-1}, \varepsilon_j\}, E \in U$, a given structure $C_E$ within $E$ is defined as:
$$C_E \subseteq \varepsilon^j = \varepsilon_1 \times \varepsilon_2 \times \cdots \times \varepsilon_{j-1} \times \varepsilon_j$$
When expressed as $E(C_E)$, this notation indicates that the existence $E$ possesses the structure $C_E$. For example, the universe's structure $C_U$ derived from $U = \{v_1, v_2, \cdots, v_{n-1}, v_n\}$ can be described as $C_U \subseteq v^n$. Similarly, Earth's structure $C_{\text{Earth}}$, derived from the existence $E_{\text{Earth}} \coloneqq \{\alpha_1, \alpha_2, \cdots, \alpha_{e-1}, \alpha_e\}$, can be expressed as $C_{\text{Earth}} \subseteq \alpha^e$. These structures $C$ are determined by the states of their respective elements.

Lastly, we formalize the concept of existences external to a given existence $E$, and the interactions between $E$ and such external existences. This formalization is essential for representing changes in the elements of an intelligent existence that occur during temporal evolution. We denote the set external to existence $E$ in $U$ as $O = \{o_1, o_2, \cdots, o_{k-1}, o_k\}, O \in U$. Formally, this is defined as:
$$O \coloneqq U \setminus E,$$
$$U = E \cup O,$$
$$E \cap O = \emptyset.$$
Here, $O$ can be interpreted subjectively as other existences external to $E$. Moreover, every existence (except for the universe itself) should engage in interactions with external existences. Intelligence, for instance, can receive input precisely because it can receive material or information from external existences, and it can produce output because it can transfer material or information to external existences. Such actions constitute interactions between existence and external existences. We formally describe these interactions, i.e., exchanges of materials or information, as occurring through temporal evolution. Specifically, we express these interactions by how the mappings $T$ project and consequently alter $E$ and $O$. At a given time $i$, the temporal mapping $T_i$ projects $E_i$ and $O_i$ onto $E_{i+1}$ and $O_{i+1}$:
$$T_i : E_i \cup O_i \to E_{i+1} \cup O_{i+1}$$
with $U_{i+1} = E_{i+1} \cup O_{i+1}$ and the cardinal condition $|U_i| = |U_{i+1}|$. If materials or information are transferred from $O_i$ to $E_i$ between times $i$ and $i + 1$, the following conditions must be satisfied:

$$|E_{i+1}| > |E_i|, |O_{i+1}| < |O_i|$$

Given the cardinality preservation $|U_i| = |U_{i+1}|$, the law of conservation of cardinality across time steps holds as follows:

$$|E_{i+1}| + |O_{i+1}| = |E_i| + |O_i|$$

Conversely, the transfer of materials or information from $E_i$ to $O_i$ would invert these inequalities. The transfer of materials or information corresponds precisely to the movement of elements from the set $O = \{o_1, o_2, \cdots, o_{k-1}, o_k\}$ to the set $E = \{\varepsilon_1, \varepsilon_2, \cdots, \varepsilon_{j-1}, \varepsilon_j\}$. If we denote the set of transferred elements as $P = \{\varpi_1, \varpi_2, \cdots, \varpi_{h-1}, \varpi_h\}$, $P \in O$, then the transition can be expressed as:

$$E_{i+1} = E_i \cup P_i,$$
$$O_{i+1} = O_i \setminus P_i$$

Here, $P_i \in O_i$ represents the set of transferred elements at time $i$. The structures of $E$ and $O$ will also change accordingly due to this transfer. Thus, the structures $C_{E,i}$ and $C_{O,i}$ evolve from time $i$ to time $i+1$ as follows:

$$C_{E,i+1} \subseteq C_{E,i} \times C_{P,i} = (\varepsilon_1 \times \varepsilon_2 \times \cdots \times \varepsilon_{j-1} \times \varepsilon_j) \times (\varpi_1 \times \varpi_2 \times \cdots \times \varpi_{h-1} \times \varpi_h),$$
$$C_{O,i+1} \subseteq C_{O,i}/C_{P,i} = (o_1 \times o_2 \times \cdots \times o_{k-1} \times o_k)/(\varpi_1 \times \varpi_2 \times \cdots \times \varpi_{h-1} \times \varpi_h)$$

Here, the symbol / denotes the removal of a particular subset Cartesian product from a Cartesian product set. Specifically, for a Cartesian product set $C = b_1 \times b_2 \times \cdots \times b_{n-1} \times b_n$ and a subset $C_p \subset C$, we define:

$$C/C_p := \prod_{b_i \in C, b_i \notin C_p} b_i$$

This formalization of interactions is equally useful for describing the transportation or interaction of materials and information within existence itself. For instance, defining subsets $E_a \in E, E_b \in E$ and describing interactions between them constitutes a formalization of internal interactions within $E$.

**Selection of a Naïve Definition of Intelligence**

Next, as a preliminary step toward an axiomatic definition of intelligence, we select a naïve definition. In this study, "intelligence" is defined as follows:

"An entity possessing structures for receiving information or matter from the external environment (input structure), processing it internally (processing structure), and outputting it externally (output structure)."

For example, a typical "neural network" fits this definition as it comprises an input layer (which receives information externally), intermediate (processing) layers, and an output layer, thus qualifying as intelligence. In contrast, a single hydrogen atom, although possessing structures such as electrons and a nucleus, cannot be considered intelligence by this definition, as these structures do not correspond to input, processing, or output functionalities. (However, as mentioned later, one could interpret even a single hydrogen atom as intelligence under an expansive interpretation; yet such an

expansive interpretation is not practically proper and thus irrelevant for intelligence creation.)

Below, we elaborate on the rationale for selecting this naïve definition. Firstly, this definition adopts a relatively broad interpretation of intelligence. It does not differentiate between living and non-living entities; it requires input, processing, and output capabilities regarding information or matter. This breadth is intentional, reflecting that higher-order intelligence is composed of combinations of lower-order intelligence [e.g., 13]. For instance, high-level intelligence, such as language recognition, emerges through combinations of simpler functions, like reflexes. Consequently, these simple functions, themselves as intelligence, could be more useful practically when creating intelligence. One concern with this definitional approach is that entities intuitively recognized by most as clearly non-intelligent could nonetheless qualify as intelligent under specific interpretations. For example, consider a "pile of sand," an entity regarded as non-intelligent. However, applying the above definition, one might state:

"A pile of sand has a structure that receives inputs (e.g., wind), changes its internal arrangement, and outputs some sand externally. Therefore, the pile of sand is intelligence."

As shown, broadly defined naïve definitions inherently bear such flaws. Moreover, attempting to reformulate this naïve definition into an axiomatic one does not resolve the issue, as will become clear in later sections. However, we consider this concern practically negligible. The reason is that when creating intelligence, we naturally avoid entities like a "pile of sand," which are useless for intelligence creation. Alternatively, defining "intelligence" itself might inherently possess little practical significance in the first place. Our goal is to create something recognized as intelligence; indeed, we are doing so. Rather than broadly outlining a conceptual boundary between intelligence and non-intelligence, as in this section, we should focus on how actual intelligence is represented within this broad framework. In other words, the definition of intelligence should inherently be constrained by practical utility. There is no meaning or necessity in defining a type of "intelligence" that is not practically useful. From this viewpoint, adopting a broadly naïve definition of intelligence—despite its intuitive flaws—is acceptable as a foundation for constructing an axiomatic definition of intelligence.

## Axiomatic Formalization of the Naïve Definition of Intelligence Based on the Set-Theoretic Universal Depiction

In this section, we formalize the previously stated naïve definition of intelligence by expressing it within the set-theoretic universal depiction described earlier. First, we define the intelligence set as $I \coloneqq \{v_1, v_2, \cdots, v_{l-1}, v_l\}, I \in U$. We express each component of the naïve intelligence definition—"an entity possessing structures to input information or matter from the external environment, process it internally, and output it externally"—using this set-theoretic framework:

**Def 1.** "Information or matter" $\coloneqq v \in U$.

**Def 2.** "Structure for externally inputting information or matter" $\coloneqq C_i \in C_I \subseteq v^l = v_1 \times v_2 \times \cdots \times v_{l-1} \times v_l$, where $C_I$ represents all structures of $I$.

**Def 3.** "Structure for processing information or matter" $\coloneqq C_p \in C_I$.

**Def 4.** "Structure for externally outputting information or matter" $\coloneqq C_o \in C_I$.

From these definitions, the entity possessing these input, processing, and output structures for information or matter is denoted as $I(C_i, C_p, C_o)$.

Next, we specify the conditions under which the set $I$ possesses the structures $C_i, C_p, C_o$. According to our earlier naïve definition, to qualify as an intelligence set, the set must have these structures. Thus, the intelligence set $I$ must have structures $C_i, C_p, C_o$, and specifying the conditions for this qualification constitutes the formal definition of intelligence in this research.

The intelligence set I and the external set $O \coloneqq U \setminus I$ evolve temporally from a certain time $i$ to time $i + 1$ through the temporal mapping $T_i$:

$$T_i: I_i \cup O_i \to I_{i+1} \cup O_{i+1}, I \in U, O \in U$$

with cardinality conservation:

$$|I_{i+1}| + |O_{i+1}| = |I_i| + |O_i|$$

Since $C_i$ is a structure for externally inputting information or matter, the condition for $I$ to have structure $C_i$ is formally defined by the following inequalities:

$$|I_{i+1}| > |I_i|, |O_{i+1}| < |O_i|$$

This change in cardinality must result from the transfer of some subset $P \in O$ into $I$.

Similarly, the condition for structure $C_O$ (external output structure) is formalized as follows:

$$|I_{i+1}| < |I_i|, |O_{i+1}| > |O_i|$$

This cardinality changes results from transferring some subset $Q \in I$ into $O$.

However, since cardinality changes for any existence between two time points can only be positive, negative, or neutral, contradictions arise if input and output occur simultaneously according to these simple conditions. To resolve this, conditions for $I(C_i, C_o)$ are defined more strictly in terms of cardinality changes between subsets of I and the external set $O$:

Specifically, we define subsets $R, S \subseteq I$:

Input condition:
$$|R_{i+1}| > |R_i|, |O_{i+1}| < |O_i|, R \in I$$
Output condition simultaneously:
$$|S_{i+1}| < |S_i|, |O_{i+1}| > |O_i|, S \in I$$

We assume that the same elements of $I$ and $O$ are not simultaneously input and output within the same time interval. Thus, defining input-output structures in this way avoids contradictions. For instance, if subsets $R$ and $S$ each have only one element transferred during a given time interval, that element is either input or output, not both. If there are suitable cardinality changes between subsets $R, S \subseteq I$ and $O$, we consider $I$ as having the structures $C_i, C_o$.

Lastly, the condition for I to possess the processing structure $C_p$ is defined by changes occurring within subsets $T, V \subseteq I$:

Internal increase condition:
$$|T_{i+1}| > |T_i|, |V_{i+1}| < |V_i|, T \in I, V \in I$$
Or internal decrease condition:
$$|T_{i+1}| < |T_i|, |V_{i+1}| > |V_i|$$

As additional subsets other than $T$ and $V$ can exist within $I$, cardinality conservation does not necessarily hold between these subsets individually. These changes are not due to element transfers with the external set $O$; instead, they reflect reorganization or restructuring of elements internally within the intelligence set. Thus, changes between time points are attributed exclusively to the internal structure of $I$. With this definition, processing structure $C_p$ formally corresponds to internal activities within an intelligence, encompassing behaviors such as distribution, transformation, memory, and restructuring of information or matter.

**Formalization and Reinterpretation of Specific Intelligences Using the Axiomatic Definition of Intelligence**

In this section, we apply the previously developed axiomatic definition of intelligence to actual intelligence examples and reinterpret their functions and mechanisms. Specifically, we examine Hebbian-learning-based non-optimized neural networks (NNs) for MNIST digit recognition [13,21], backpropagation-based optimized NNs for MNIST digit recognition [e.g., 19,21], and the gill-withdrawal reflex of Aplysia [e.g., 20]. By doing so, we discuss the differences among these intelligences, their levels of biological plausibility, and constraints or prohibitions in creating biologically inspired artificial intelligences. First, we summarize their respective sets, elements, and structures in Table 1. Then, we detail how inputs, processing, and outputs occur over time in Table 2. In Table 2, the operational processes are divided into separate stages of learning and testing or actual operations, with input, processing, and output assumed to occur sequentially across distinct time steps.

|  | **Hebbian Non-optimized NNs (MNIST classification) [13]** | **Backpropagation Optimized NNs (MNIST classification) [e.g., 19]** | **Aplysia Gill-Withdrawal Mechanism [e.g., 20]** |
|---|---|---|---|
| **Set Classification** | Computer and program code defining NNs and related functions. Ten parallel NNs. | Computer and program code defining NNs and related functions. Non-parallel NNs. | Cells are involved in gill movement within Aplysia's body. Non-parallel neural system. |
| **Elements** | Computing devices within computers. Virtually: Nodes defined by program code and their connections, Hebbian weight update structures, vector-norm calculation structures. | Computing devices within computers. Virtually: Nodes defined by program code and their connections, plus loss function, backpropagation-based weight updating functions, softmax functions. | Individual biological cells within Aplysia, including neural cells. Specifically, it includes sensory receptor cells, motor neurons, muscle tissues, and related neurotransmitters. |
| **Structure $C_i$** | Input layer. | Input layer, loss function, output layer. | Structure for external stimulus reception via surface sensory receptors (e.g., touch or water flow detection). |
| **Structure $C_p$** | Intermediate (hidden) layers. | Intermediate (hidden) layers, backpropagation functions. | Internal neural structures are maintained by nerve cells. Neurotransmitter production and release functions. Hebbian synaptic reinforcement. |
| **Structure $C_o$** | Output layer. Vector-norm calculation for correct/error determination. | Output layer. Softmax function. Node count adjusted according to labels. | Gill-opening and closing mechanism via muscle contraction and relaxation. Threshold-based structure. |

Table 1: Summary of Intelligence Sets, Elements, and Structures

| Learning Phase Time Mapping | Hebbian Non-optimized NNs [13] | Backpropagation Optimized NNs [e.g., 19] | Aplysia Gill-Withdrawal Mechanism [e.g., 20] |
|---|---|---|---|
| $T_i: I_i \xrightarrow{c_i} I_{i+1}$ | Input of label-specific training data into parallel NNs' input layers. Response to intermediate layers. | Input of training data into input layer. Response to intermediate layers. | Detection of physical stimuli by sensory receptor cells. Response to neural cells. |
| $T_i: I_{i+1} \xrightarrow{c_p} I_{i+2}$ | Forward propagation through intermediate layers. Hebbian weight updates. Response to output layer. | Forward propagation through intermediate layers. Response to output layer. | Forward propagation of signals among neural cells. Hebbian reinforcement of synaptic connections. Includes short-term and long-term learning, habituation. |
| $T_i: I_{i+2} \xrightarrow{c_o} I_{i+3}$ | Response of output layer (no meaningful response in Itoh, 2024b). | Response of output layer to loss function. | Gill-opening/closing response based on unlearned decision criteria (threshold). |
| $T_i: I_{i+3} \xrightarrow{c_i} I_{i+4}$ | - | Response of loss function to output layer. | - |
| $T_i: I_{i+4} \xrightarrow{c_p} I_{i+5}$ | - | Backpropagation in intermediate layers and weight updates. | - |
| **Testing/Operation Phase Time Mapping** | **Hebbian Non-optimized NNs [13]** | **Backpropagation Optimized NNs [e.g., 19]** | **Aplysia Gill-Withdrawal Mechanism [e.g., 20]** |
| $T_i: I_i \xrightarrow{c_i} I_{i+1}$ | Input of test data into input layers. Response to intermediate layers. | Input of test data into input layer. Response to intermediate layers. | Detection of physical stimuli by sensory receptor cells. Response to neural cells. |
| $T_i: I_{i+1} \xrightarrow{c_p} I_{i+2}$ | Forward propagation in intermediate layers. Response to output layer. | Forward propagation in intermediate layers. Response to output layer. | Forward propagation of signals among neural cells. |

| $T_i: I_{i+2} \xrightarrow{C_o} I_{i+3}$ | Response in output layer. Correct/error determination based on vector-norm calculation. | Response in output layer. Correct/error determination. | Gill-opening/closing response based on learned decision criteria (threshold). |

Table 2: Summary of Set Changes with Temporal Evolution of Intelligence Sets. The symbol $\xrightarrow{C}$ indicates that a structure $C$ contributes to set changes during temporal evolution.

As shown in Tables 1 and 2, Hebbian-learning-based non-optimized NNs possess a local Hebbian weight-update structure and vector-norm calculation structure, functioning as both learning and error-determining structures. Consequently, their learning processes rely solely on local weight updates without a loss function. Conversely, backpropagation-based optimized NNs update weights based on a non-local loss function and backpropagation mechanism, typically employing softmax functions for their output structure. In contrast, the biological intelligence represented by Aplysia's gill-withdrawal reflex involves interactions among neural populations and muscle groups, generating outputs as gill movements based on internal threshold criteria. An analysis of their temporal evolution processes in Table 2 reveals apparent similarities between Hebbian-learning-based NNs and Aplysia's neural circuitry. Both exhibit local, self-organizing structural changes in neural systems in response to external inputs, enabling learning without reliance on non-local error functions. However, Backpropagation-based optimized NNs require external loss functions, resulting in more necessary time steps in their learning process compared to the other two systems. Thus, backpropagation-based optimized NNs significantly diverge from biological plausibility.

However, even Hebbian-learning-based non-optimized NNs contain certain non-biological features. The model presented by Itoh (2024b) [13] employs parallel NN structures, with each network trained independently for specific labels. While such parallel architectures offer practical engineering benefits for easily demonstrating Hebbian utility, they significantly diverge from biological neural systems' non-parallel, integrated information processing characteristics. Typically, biological nervous systems handle multiple functions within single neural structures, rendering a parallel, label-specific structure artificial and non-biological. Furthermore, although the Hebbian learning rule itself may be biologically plausible, the objective for which it is used—such as high-level functions like digit recognition—is fundamentally distinct from biological reflexive behavioral mechanisms. Biological reactions, such as Aplysia's gill withdrawal, are low-level reflex systems eliciting specific physical responses to specific stimuli, involving no abstract informational processing like "classification" or "recognition." Consequently, designing simple structures aiming at digit recognition implicitly assumes a fundamentally non-biological processing objective even if labeled biologically inspired. Hence, implementing biologically plausible digit recognition would require genuinely biological mechanisms for recognition tasks.

In summary, Hebbian-learning-based non-optimized NNs possess biological characteristics regarding local learning and non-optimized structures. Nonetheless, their parallel architecture and functional objectives introduce explicit non-biological elements. Therefore, it is insufficient to merely incorporate Hebbian learning rules to enhance biological plausibility and ultimately create biologically inspired AI. It is necessary to re-examine their application methods, structural designs, and even intended functional goals. This analysis suggests that non-local loss functions or backpropagation mechanisms should be considered prohibitive in biologically inspired AI design. In contrast, local, self-organizing Hebbian-like learning structures should be actively pursued. However, careful attention must also be given to other non-biological elements, such as parallel NN structures, which might be restricted depending on the desired application. Additionally, progress in developing human-inspired AI or AGI would benefit significantly from extending such axiomatic understanding of intelligence toward humans and AGI. This research could serve as a foundational theory toward that goal.

As an additional discussion, it is worthwhile to consider more advanced axiomatic definitions of intelligence and their formal descriptions. Throughout this study, we consistently sought to reinterpret naïve intelligence definitions more axiomatic. While this research provided formal frameworks labeled as "intelligence," the specific contents and operational details of intelligence have been mainly described through natural language. In the future, complete axiomatic formalization of specific intelligence—including their detailed contents and operational dynamics—should be pursued. Achieving this would facilitate more axiomatic understandings of intelligence and enable circuit-level descriptions of intelligence and potentially their hardware implementations.

**Discussions**
**The Arbitrariness of the Axiomatic Intelligence Definition and Its Exemplification via Extension to Category Theory**

In this study, we developed an axiomatic definition of intelligence through a set-theoretic formalization. However, it is important to acknowledge the inherent arbitrariness of the methodology used for this axiomatic definition. The definitions and methodologies proposed here represent just one possible approach. Alternative theories—potentially superior in certain aspects—or entirely different mathematical frameworks and choices of naïve definitions could also be developed. The necessity for such alternatives is guided by the usefulness of the definition itself, as outlined in the research objectives section. To illustrate this arbitrariness, we briefly describe how intelligence might alternatively be defined using category theory rather than set theory. Accepting the set-theoretic universal depiction, as well as its definitions of existence, structure, and naïve definition of intelligence, we introduce a more advanced and higher-order structural formalization of intelligence and its temporal evolution through morphisms and functors between two categories: the Intelligence Category $\mathcal{I}$ and the Time Category $\mathcal{T}$. Intelligence, as an object possessing input, processing, and output structures, can be formalized categorically as follows:

$$\mathcal{I} = \text{Cat}_{\text{Intelligence}}, \text{Obj}(\mathcal{I}) = \{I = (C_i, C_p, C_o)\}$$

$$\text{Mor}_{\mathcal{I}}(I, J) = \{f: I \to J | f = (f_i, f_p, f_o), f_k: C_k^I \to C_k^J, k \in \{i, p, o\}\}$$

Here, the object $I$ denotes a specific intelligence, represented by the input set $C_i$, processing set $C_p$, and output set $C_o$. Additionally, $\mathcal{I}$ represents all structurally defined intelligence objects $(C_i, C_p, C_o)$. A morphism $f: I \to J$ represents a structure-preserving map from one intelligence to another, encapsulating transformations such as learning, conversion, or imitation, with individual component mappings $f_k$.

On the other hand, the temporal evolution is defined as a separate category—the Time Category $\mathcal{T}$—where time points are objects, and the progression of time is represented by morphisms:

$$\mathcal{T} = \text{Cat}_{\text{Time}}, \text{Obj}(\mathcal{T}) = \{t_i\}$$

$$\text{Mor}_{\mathcal{T}}(t_i, t_j) = \begin{cases} \{\tau_{ij}\} & \text{if } i \leq j \\ \emptyset & \text{otherwise} \end{cases}$$

Here, the morphism $\tau_{ij}: t_i \to t_j$ expresses the temporal transition from time $t_i$ to $t_j$, satisfying temporal continuity via the composition $\tau_{ij} \circ \tau_{jk} = \tau_{ik}$.

The temporal evolution of intelligence can then be formalized through a time functor $F_I: \mathcal{T} \to \mathcal{I}$:

- For each time $t_i \in \text{Obj}(\mathcal{T})$, we associate a corresponding intelligence object $I_i = F_I(t_i) \in \text{Obj}(\mathcal{I})$.
- For each morphism $\tau_{ij}: t_i \to t_j \in \text{Mor}(\mathcal{T})$, we associate a structural transformation of intelligence $f_{ij} = F_I(\tau_{ij}): I_i \to I_j \in \text{Mor}(\mathcal{I})$.

Thus, the functor $F_I$ satisfies the following conditions:

- Preservation of identities: $F_I(\text{id}_{t_i}) = \text{id}_{I_i}$
- Preservation of compositions: $F_I(\tau_{jk} \circ \tau_{ij}) = F_I(\tau_{jk}) \circ F_I(\tau_{ij})$

Different intelligence objects may correspondingly be associated with distinct time functors $F: \mathcal{T} \to \mathcal{I}$. Thus, the temporal evolution of each intelligence is consistently represented by its corresponding time functor $F$. Additionally, because the morphism $f_{ij} = F(\tau_{ij})$ resides within the Intelligence Category, state changes at each time step may be generalized to include other transformations, such as learning processes, defined as morphisms.

Moreover, if multiple intelligence systems (for instance, biological and artificial intelligences) are defined as distinct Intelligence Categories $\mathcal{I}_{\text{Bio}}, \mathcal{I}_{\text{AI}}$, the relationships between them—such as artificial imitation of biological intelligence—can be mathematically formalized via inter-intelligence-category functors $G: \mathcal{I}_{\text{Bio}} \to \mathcal{I}_{\text{AI}}$. For example, Table 1 could be described using the Intelligence Category $\mathcal{I}$, its morphisms, and functors $G$, while Table 2 could be expressed using the Intelligence Category $\mathcal{I}$, Time Category $\mathcal{T}$, their morphisms, and functors $F, G$, or their composite functor $H = F \circ G$. Thus, this approach can serve as a methodological foundation for achieving complete axiomatic intelligence formalization as discussed earlier. Furthermore, morphisms within the Intelligence Category and functors between intelligence categories may represent not only temporal transformations—such as learning-induced changes—but also non-temporal functional transformations like imitation, translation, and abstraction. Compositions of such transformations could collectively represent large-scale and higher-order structural mappings, for example, from biological intelligence to fully biologically inspired artificial intelligence. While the set-theoretic framework emphasizes concrete element movements, the categorical framework focuses more on how functions within intelligence correspond and can be transformed. Implementing axiomatic intelligence definitions through such diverse conceptual frameworks would deepen our understanding of intelligence significantly and greatly contribute to the development of human-inspired AI and AGI.

**On the *Activity* of Intelligence**

The axiomatic definition of intelligence proposed in this study requires that intelligence be defined as a set that engages in inter-set interactions through temporal evolution. Therefore, any set that does not interact with other sets during time evolution cannot be regarded as an intelligence set at that moment—even if it structurally possesses input, processing, and output capabilities. In other words, even if a set has an internal structure corresponding to intelligence, it does not function as intelligence during any temporal interval in which its structure is not actively operating due to some external or internal cause. For instance, in the case of artificial intelligence, even if a system contains structures that fulfill the intelligence criteria, it cannot be considered active intelligence if the computer is not powered on. No meaningful interaction with the external world occurs. In such a state, intelligence does not manifest, thus failing to qualify as intelligence under the current definition. This implies that the definition implicitly assumes specific material and environmental conditions must be met for intelligence to function. If we wish to treat this point more rigorously, then in addition to defining the structure of intelligence, the conditions under which that structure functions as intelligence should also be explicitly incorporated into the definition. This perspective becomes clearer when one introduces a property called the *activity* of intelligence. Here, *activity* refers to a quantitative measure indicating the degree to which a given intelligence structure functions as intelligence over a particular time interval—namely, the extent to which it engages in interactions with other sets. For example, an intelligence defined by some program structure will not perform any intelligent function if it is not actively running. Hence, an artificial intelligence system during a period in which it is not activated may be evaluated as having zero intelligence activity. In contrast, living beings engage in continuous interactions with the external world throughout their life span, from birth to death, and can thus be considered to exhibit high intelligence activity. Of course, there may be differences in activity between states, such as sleep and wakefulness. In this way, the activity could serve as a valuable auxiliary concept for defining and distinguishing different types of intelligence, for instance, biological vs. non-biological intelligence, human vs. non-human intelligence, or AGI vs. non-AGI.

**Axiomatic Definitions of Concepts Other Than Intelligence**

The methodology used in this study to axiomatize intelligence should also apply to concepts other than intelligence. That is, concepts such as *consciousness* or *emotion*, which are considered essential to advanced intelligence development and which—like intelligence—are originally naïve, can be axiomatized through the same procedure:

Definition of a set-theoretic universal depiction → Selection of a naïve concept → Axiomatic formalization of the naïve concept within the universal depiction

These concepts might be positioned as partial functions within the broader definition of intelligence, or they may be treated as separate but parallel constructs. As discussed throughout this study, it is methodologically acceptable to maintain multiple definitions, and the framework is flexible enough to permit that. These definitions may be formulated using set theory, or category theory may be employed as in the previous section. Other theoretical frameworks—such as topos theory, for example—could also be used. Moreover, if additional concepts beyond "consciousness" and "emotion" are necessary and valuable, they, too, should be axiomatized accordingly. This study may be interpreted as offering a methodology for treating metaphysical concepts—traditionally handled in philosophical discourse—as physical or formalizable phenomena. In other words, this may be an attempt to transform a part of what Kant referred to as the "thing-in-itself (*Ding an sich*)" [22] and what Wittgenstein described as "Whereof one cannot speak, thereof one must be silent. (*Wovon man nicht sprechen kann, darüber muss man schweigen.*)" [23] into "speakable phenomena."

# Toward the Axiomatization of Intelligence: Structure, Time, and Existence

伊藤　慧*

April 20, 2025

## Abstract

本研究は、知能という本質的に素朴かつ多義的な概念に対して、その定義方法自体を定義するというメタ的枠組みのもと、公理的定義を構築することを目的とする。まず、知能が存在する場としての宇宙を集合論的に形式化し、その中で時間発展や他集合との相互作用を考慮した構造として知能を記述する。次に、"情報や物質を、外部から入力・内部で処理・外部へ出力する構造を有する存在"とする素朴的知能定義を出発点とし、それを集合論的宇宙描像のもとで公理的に定式化する。この公理的知能定義を適用し、ヘブ則型非最適化ニューラルネットワーク（NNs）、逆伝播型最適化 NNs、および生物の反射行動系（アメフラシの鰓開閉）という三例において知能やその構造、そしてその生物模倣性を比較・解釈する。さらに、この知能定義を圏論の枠組みに拡張し、時間圏と知能圏という二つの圏とそれらの間の関手構造を導入することで、知能の変化や模倣関係をより抽象的に表現しうる可能性を示す。また、本定義により示される知能は時間的な相互作用を伴ってはじめて知能として機能するという観点から、活動度と呼称する概念を導入し、その活動度に基づいて知能の活動条件によるその知能らしさや知能分類を考察する。最後に、本研究で提示した定義手法は知能のみに限らず、意識や感情といった他の概念にも適用可能であることを示唆し、これらを同様の手順——宇宙描像の定義、素朴的定義の選択、公理的定義化——によって形式的に再解釈することが可能であると論じる。以上より、本研究は、知能定義の基礎理論としての有用性にとどまらず、様々な従来形而上的とされてきた概念の形式的記述のための一般的な枠組みとしても重要な示唆とその定義のための手法を提供しうる。

\* Ehime University Graduate School of Science and Technology. The author conducted this research independently while affiliated with Ehime University, and utilized the university's library and electronic resources. The research was conducted without specific financial support from external funding sources. The author declares no conflict of interest.

## 知能の定義の素朴さとその例

　　　　我々人間は"知能"を持つ。また、人間以外の生物やコンピュータープログラムで構成された人工知能、果てはより単純な道具にもその内に知能が宿っているか、もしくはそう見える。この"知能"というもの、もしくはそれに類似した概念をどのように定義するかという課題について古代から現代にいたるまで様々な人々が取り組んできた。例えばプラトンは"国家"第6巻（例えば509d–511e）[1]において、"真なる知能"（*νοῦς*）を、仮定に依拠する数学的推論（*διάνοια*）のための能力ではなく、真理やイデアを無仮定に把握する思考（*νόησις*）を可能とする能力であるとしている。カントは *Verstand*（悟性）を"*Verstand ist aber das Vermögen der Begriffe.*（悟性とは概念の能力である）"（A69/B94）[2]とし、これは与えられた直観に対して判断・統合を行う知的機能であると位置付けている。この点において *Verstand* は現代的な"知能"の主要な構成要素と見なすことができ、したがって知能を定義づける概念に他ならない。また、心理学の観点からウェクスラーは"*The aggregate or global capacity of the individual to act purposefully, to think rationally, and to deal effectively with his environment.*（個人が意図的に行動し、合理的に考え、環境に効果的に対処する能力の総体または全体。）"[3]と知能を定義した。あるいは計算学・論理学的観点からチューリングは"知能を持つかどうかを判定する"ためにチューリングテストを考案した[4]。これは知能の直接的な定義を避けているがしかし、間接的に知能の定義を行うという意味で、やはり知能の定義に他ならない。加えて、シュレディンガーは知能というよりも生命の性質について"*It is by avoiding the rapid decay into the inert state of "equilibrium" that an organism appears so enigmatic... What an organism feeds upon is negative entropy.*（平衡状態の急速な崩壊を回避することで、生物は不可解な存在となる。生物が何を餌としているかというと、それは負のエントロピーである。）"[5]と述べた。

　　　　これらの他にも幾百の"知能の定義"が存在しているわけだが、つまりこのことは知能というものが素朴的（非公理的）定義によって定義付けられていて、そして一義に定義されないものであるということを示している。そして一義に定義されていないことは人々・社会の間で"知能"について統一された概念が存在しないということでもある。例示した5つの定義や解釈を比べても、一部分は対応しているように見えたとしてもそれぞれが完全に対応するものだとは言えない。プラトン、カント、ウェクスラー、チューリング、そしてシュレディンガーにとっての"知能"が何であるかというのは当人たちの認識からしてそれぞれに違うのであって、それは他の全ての人間でも同様であるに違いない。こうした知能の定義の多様性については、AI理論の分野でも深く認識されており、Legg & Hutter (2007) [6]は70以上の知能定義が文献上に存在することを確認しつつ、それらを代表する定義群を収集・整理している。このように、知能とは何かという問題は、哲学や心理学に限らず、人工知能の理論枠組みにおいても未解決かつ根本的な課題であるといえる。

**AI 開発における知能やその能力の公理的定義や数学的形式化の重要性**

　　　　他方、人工知能の制作を含め、知能を作るという営みもまた人類の歴史（あるいは生命の歴史）において行われ続けてきた。そして収穫加速の法則[7]によれば人類（生命）によって作られる知能とは指数関数的に進歩していく。このことは今現在人類が作成しようとしている知能（例えば Artificial general intelligence (AGI)など）とは指数関数的にそのレベルや複雑さ、あるいは作成難度を増加させていることを同時に示すだろう。ここで"知能"が素朴的に定義されており、そして一義に定義されていないことは、その高度な知能を作成する上で様々なそして重大な問題を発生させ得る。以下にその例を二つ示す。

　　　　AGI 開発の現場では、"大規模言語モデル（LLM）を拡張すれば高い知能が獲得できる"と主張する立場[e.g., 8,9]と、"エージェントとして環境との相互作用（行動と試行錯誤）を重視しなければ知能とは呼べない"[e.g., 10,11]とする立場とのあいだで、対立が顕在化している。前者は言語処理や論理推論能力を重視するが、後者は行動適応能力こそが知能の核心だと考えている。両者ともに"知能"の一面を確かに捉えてはいるものの、知能そのものの定義が多義的で合意が得られていないがゆえに、研究のアプローチや評価基準が大きく異なり、相互理解や統合が進みにくい。これは高度な知能の開発を目指すうえでの効率低下のみならず、人工知能の社会的な実装においても基盤となるべき概念の統合の欠如につながりかねない。

　　　　Itoh(2024a, 2024b)[12,13]はニューラルネットワーク（NNs）[e.g., 14,15]-AI を用いた生物神経系の模倣のためにはバックプロパゲーションなどの誤差最適化手法を含めた非生物的構造を用いることは不適であるとした。また実際にヘブの法則[16]を NNs に実装するためには直接にベクトル類似度を計算する非生物的な分散表現を用いることは不適であることを示し、より生物的に妥当な構造を作成することでその実装を可能とした[13]。これはつまり、ある目的の知能（ここでいう生物神経系模倣 AI のような）を作る場合において"禁則事項"が存在しうるということである。しかし知能作成において知能の定義が素朴的かつ多義であると禁足事項も素朴的かつ多義であることに繋がり、知能制作者間もしくは社会間において明確かつ厳密な知能の定義と禁則事項の定義を行うことができない。そして、そのような比較的高度な定義とその社会間の統合された一致が必要な知能を作成することは不可能だということになるのである。あるいは、NNs-AI 開発理論においてヘブの法則のようなその本来的な概念ではなく最適化過程[e.g., 17]やバックプロパゲーション[e.g., 18]のような副次的な概念が主流となった要因も、知能の素朴的理解と多義性による知能やその禁則事項の公理的理解の欠如によるところが大きいだろう。このように高度な知能の作成とその社会的な実装には知能の高度な理解やそこから派生する禁則事項の理解が必須であり、そのためには知能の素朴的かつ多義な理解からの脱却が必須なのである。

### "知能の定義"をどのように定義付けするか

しかしながら、"知能"そのものが本質として素朴的な概念であり、そして多義であることもまた明らかである。知能は数学的な証明が可能なものではなく、物理学的な観測が可能なものでもなく、知能とは我々人間が素朴的に理解しそしてそれぞれに定義付けるものだからである。これはつまり、"知能の定義"が多義であるとも言えるのである。人や社会それぞれにとって知能の定義が違い、そしてその方法論ももちろん違うだろう。ある人にとっての知能の定義が例え記述していなくともあるいは他者に伝達できずとも、その人にとって知能という定義が何であるかというのは、もしもその人にとって自明ならばすなわち自明であり、そしてその"知能の定義"とは明らかに正しいのである（もしくは否定不可能なのである）。翻って、知能の素朴的かつ多義な理解からの脱却には、それが可能な"知能の定義の定義付け"が必要であるのだと私は考える。つまり、知能というものを定義するための方法を定義することから行わなければならないということである。もしもこの"知能の定義の定義"が、素朴的かつ多義な理解からの脱却が可能なだけの厳密性を持った公理的かつ一義な知能の定義を可能とするのならばすなわち、相応に高度な知能が作成可能であると言えるだろう。加えて、"知能の定義の定義"やこれによって形作られた"知能の定義"に一定の論理的・科学的な頑強さがあるのならば、知能の多義的な理解が知能制作においてもある程度許容され得るかもしれない。たとえいくつかの知能定義があってもそれらの公理性が十分に高いのならば、それらは完全な比較や対応が可能なものであるかもしれない。そしてある知能とその能力をそのいくつかの公理的知能定義を用いて多面的に解釈した時に、何かしらの利益を得られる可能性すらあるだろう。例えば後述する知能の集合論的定義と圏論的定義は互いに異なる"知能の定義"であるが、同時に互いに対応させられ得るだけの数学的厳密性を担保することが可能なはずである。本研究において集合論的描像は知能と知能外間の要素の移動に着眼し、圏論的描像は知能間の機能の対応付けに着眼し知能を解釈するが、これらの解釈は知能制作においてどちらとも必要である場面が発生するはずである。

### 宇宙の集合論的描像

我々が知能を持つと考えるおよそのものは、この宇宙の中で物質としてある構造を持ち存在している。そして当然、有知能の物質と無知能の物質の違いとは、その物質の構造が知能足りえる構造であるかどうかに起因してもいるはずだ。このことから非素朴的な知能の定義もしくは知能の定義の定義のためには、知能が存在する宇宙の描像を公理的に定義しなければならない。知能とはある宇宙の中のある物質とその構造なのであって、その内容を定義したければまずは宇宙そのものを定義しなければならない。本研究では公理的知能定義の前にその知能が存在する宇宙描像を公理的に定義し、それを基に知能の定義を行っていく。本研究ではこの描像には宇宙が粒子の集合であると公理的に定義する"集合論的宇宙描像"を基本的には採用する。またこの描像の中で、"時間"をある時刻からある次の時刻へと宇宙粒子集合を投射する写像として公理的に定義する。これらの集合論的宇宙描

像と時間写像を以てして知能と知能が持つ能力を定義・解釈する。またこの宇宙描像の定義やこのような方法を採るということ自体が"知能の定義の定義"だと言えるだろう。ここで、宇宙集合や時間写像の厳密性は定義したい知能を定義するに十分な程度に担保できていれば十分であるとする。つまり、ある知能は古典粒子あるいは古典力学の範囲で定義しうるかもしれないし、量子的粒子あるいは量子力学まで範囲を広げる必要があるかもしれないが、その定義の細かさは必要なだけ細かければ十分だろうということだ。そしてあるいは、粒子集合の解像度や力学的作用は相当に限定的な記述で十分である場合も多いだろう。例えばNNsはプログラム上に関数やアルゴリズムで定義された"知能"であって（ここでは知能であるとして）、NNsの内的存在や能力のみに限ればその記述はそのプログラムの機序と作用のみでおそらくは十分だということである。

## 研究目的

本研究では以上のことを踏まえて知能の公理的定義を行い、そしてその定義に基づいた簡単な知能の持つ能力の解釈を行う。知能の公理的定義は以下の手順で行う。
1. 集合論的宇宙描像の公理的定義と選択
2. 素朴的知能定義の選択
3. 集合論的宇宙描像に基づく素朴的知能定義の公理的定義化

2は前述の通り"知能"が元来素朴的概念であることから必要な過程である。公理的知能は公理的知能定義から直接定義されるのではなく、ある素朴的知能定義を公理的定義化することによって定義される。3では定義された公理的知能を基に具体的な知能を再解釈することで知能の持つ能力の解釈の例をも示す。具体的にはItoh(2024b)[13]によるヘブ学習型非最適化NNsと従来的な逆伝搬型最適化NNs[e.g., 19]、そしてアメフラシの鰓開閉機構[e.g., 20]の3つの"知能"に対し本研究による公理的知能定義を適用する。そしてそれらを比較することによりそれぞれの知能性やその差異について解釈を行う。最後に本研究における知能定義の任意性を集合論から圏論へ議論を簡単に拡張することで改めて確認したり、知能定義から導かれる知能の"活動度"についての記述をしたり、そして"意識"や"感情"といった他概念の形式化の可能性について記述したりすることで考察を付記する。

本論の目標は、知能とその能力の公理的定義が行えるようになることだが、そもそも"知能"という概念を定義することの直接に実用的な意味はあまり大きいと考えていない。直接に実用的に意味のある定義とはむしろ、"○○という知能"の定義だと考えている。つまり我々が作成したいのは知能という概念や機構ではなく、むしろ知能と呼ばれている何か概念や機構なのであって、そしてその公理的定義化が必要なのである。そういう意味で"知能"それ自体の定義は一つの例示でしかない。あるいは我々が作るべきものの概念や構造の最も広義な枠組みであるとも言えるだろう。

**集合論的宇宙描像の公理的定義と選択**

　　　　本セクションでは、知能が存在する宇宙およびその時間発展を、集合論的な枠組みを用いて定式化する。まず、全宇宙集合を$U$とし、これを$U$の要素$v$の全集合として表現する。すなわち、全宇宙集合$U$は以下のように定義される。

$$U \coloneqq \{v_1, v_2, \cdots, v_{n-1}, v_n\} = \{v|\text{Universe}\}$$

つまり$U$は宇宙のある要素$v$を全て集めた有限集合である。要素$v$は例えば原子であるかもしれないし素粒子かもしれないし、あるいはニューロンと捉えることができるかもしれないが、それをここで詳細に記述する必要はなく、具体的知能を定義する際に要素も同様に具体的に記述すべきである。そして$v$はそれぞれにある状態（位置や物理量など）を持つだろうし、その詳述が必要になる場合もあるだろうが、それも同様に具体的知能を定義する際に記述すればよい。本項ではあくまで知能の存在する宇宙という枠組みを定義したいのであって、加えてその宇宙の具体的記述は定義される知能により異なるべきであるはずでもあるから、要素や状態の記述については知能定義に付随して行うべきだろう。

　　　　$U$は時間写像$T$によってある時刻から次の時刻へと投射される。つまりある時刻$t_i$における$U$である$U_i$は$T_i$によって時刻$t_{i+1}$における$U$である$U_{i+1}$へと時間発展するのであり、このことは集合論形式により以下のように表現される。

$$T_i: U_i \to U_{i+1}$$

ただし$U$の濃度は一定である。よって時間の経過は

$$T_i: U_i \to U_{i+1},$$
$$T_{i+1}: U_{i+1} \to U_{i+2},$$
$$T_{i+2}: U_{i+2} \to U_{i+3},$$
$$\cdots,$$

と$T$による$U$の連続的な投射によって表現される。ここで$U$に対する記述と同様に、$T$に対する詳細の記述に関してもそれはここで定義すべきものではなく、具体的な知能定義の際に行うべきものであるとする。つまり時間の連続性や細かさをどのように考慮するかや時間発展に伴う力学的作用の記述は定義したい知能によってそれぞれに合わせるべきであり、本セクションで行うべきことではないということである。

　　　　また知能定義のためにこの宇宙におけるいくつかの概念を定義・形式化する。具体的にはこの宇宙における存在や構造という概念、加えて存在と存在外との相互作用を集合論形式から形式化する。$U$における存在$E$とは、$U$の任意の部分集合であると定義する。つまり$U$における存在$E$は

$$E \subseteq U.$$

例えば$E = U$であるときは$E$は宇宙そのものという存在であると考えられる。また例えば、$E_{\text{Earth}}$を地球という存在だとすれば$E_{\text{Earth}} \coloneqq \{\alpha_1, \alpha_2, \cdots, \alpha_{e-1}, \alpha_e\} = \{\alpha|\text{Earth}\}(\text{Earth} \in \text{Universe})$だと形式化できるし、ここで地球は明らかに宇宙全体と同一の存在ではないから$E_{\text{Earth}} \subset U$である。ただ地球（あるいはそれ以外のある存在）の要素全ての集合を規定する

方法は要素や時間と同様に知能定義の際に詳述されるべきことだろう。

　　　　　ある存在はある構造$C$を有し、$C$とは存在における要素同士のある関係（相対位置など）であって、以下のように直積集合を用いて形式化される。$E = \{\varepsilon_1, \varepsilon_2, \cdots, \varepsilon_{j-1}, \varepsilon_j\}, E \in U$として、$E$におけるある構造$C_E$は

$$C_E \subseteq \varepsilon^j = \varepsilon_1 \times \varepsilon_2 \times \cdots \times \varepsilon_{j-1} \times \varepsilon_j$$

と定義する。そして$E(C_E)$と表記した場合にはその$E$はある構造$C_E$を有するとする。例えば宇宙の構造$C_U$は$U = \{v_1, v_2, \cdots, v_{n-1}, v_n\}$から$C_U \subseteq v^n$と表現できるし、地球の構造$C_{\text{Earth}}$は$E_{\text{Earth}} \coloneqq \{\alpha_1, \alpha_2, \cdots, \alpha_{e-1}, \alpha_e\}$から$C_{\text{Earth}} \subseteq \alpha^e$のように表現できるだろう。これらの$C$は要素のそれぞれの状態によって決定されるはずである。

　　　　　最後にある$E$外の存在やその$E$との相互作用について記述する。これは時間発展に伴う知能存在の要素の増減の表記に重要な形式化である。$U$においてある存在$E$ではない集合を$O = \{o_1, o_2, \cdots, o_{k-1}, o_k\}, O \in U$と表現する。つまり、

$$O \coloneqq U \setminus E,$$
$$U = E \cup O,$$
$$E \cap O = \emptyset.$$

ここでの$O$は$E$にとっての主観としての他者存在であると解釈可能だろう。また（$U$自身を除いて）どのような存在も他者存在との相互作用を行っているはずである。特にある知能がある入力を受けることができるのは、他者存在からの何らかの物質や情報の受取が可能であるからだし、ある出力を行うことができるのは他者存在への何らかの物質や情報の受渡が可能であるからであり、これらのことはそのある存在と他者存在との相互作用が成立していることに他ならない。ここでこのある存在とその他者存在との物質や情報の授受についての形式化を行っておく。この相互作用とは時間発展の過程において行われることは明らかであるから、$T$による$E$や$O$の投射によってそれらがどのように変化するかを記述することで表現する。時刻$i$における時間写像$T_i$は同じく$E_i$や$O_i$を時刻$i+1$における$E_{i+1}$や$O_{i+1}$へ投射する。つまり、

$$T_i : E_i \cup O_i \to E_{i+1} \cup O_{i+1}.$$

ただし$U_{i+1} = E_{i+1} \cup O_{i+1}$であり、$|U_i| = |U_{i+1}|$とする。ここで例えば、時刻$i$から$i+1$にかけて$O_i$から$E_i$に対して物質や情報が受け渡されたということは以下の条件を満たすことになる。

$$|E_{i+1}| > |E_i|, |O_{i+1}| < |O_i|.$$

ここで$|U_i| = |U_{i+1}|$から、

$$|E_{i+1}| + |O_{i+1}| = |E_i| + |O_i|$$

という時刻間の濃度の保存則が成り立つ。また$E_i$から$O_i$への物質や情報の受け渡しであれば符号を反転し、$|E_{i+1}| > |E_i|, |O_{i+1}| < |O_i|$と表現されることは明らかである。ここでこのような物質や情報の受け渡しは$O = \{o_1, o_2, \cdots, o_{k-1}, o_k\}$のある要素が$E = \{\varepsilon_1, \varepsilon_2, \cdots, \varepsilon_{j-1}, \varepsilon_j\}$へ移動することに他ならず、このことは以下のように表現できる。受け渡される要素の集合を

$P = \{\varpi_1, \varpi_2, \cdots, \varpi_{h-1}, \varpi_h\}$, $P \in O$ として、
$$E_{i+1} = E_i \cup P_i,$$
$$O_{i+1} = O_i \setminus P_i.$$
ここで$P_i$は時刻$i$における$P$であり、$P_i \in O_i$である。そしてこのような物質や情報の移動により$E$や$O$の構造も変化するだろう。つまり$E_i$の構造$C_{E,i}$、$O_i$の構造$C_{O,i}$は時刻$i$から時刻$i+1$に時間発展する中で以下のように変化する。
$$C_{E,i+1} \subseteq C_{E,i} \times C_{P,i} = (\varepsilon_1 \times \varepsilon_2 \times \cdots \times \varepsilon_{j-1} \times \varepsilon_j) \times (\varpi_1 \times \varpi_2 \times \cdots \times \varpi_{h-1} \times \varpi_h),$$
$$C_{O,i+1} \subseteq C_{O,i}/C_{P,i} = (o_1 \times o_2 \times \cdots \times o_{k-1} \times o_k)/(\varpi_1 \times \varpi_2 \times \cdots \times \varpi_{h-1} \times \varpi_h).$$
ここでの/記号は以下のようにある直積集合からそのある部分直積集合を除去することを意味する記号として扱う。つまりある直積集合$C = b_1 \times b_2 \times \cdots \times b_{n-1} \times b_n$とその部分集合$C_p \subset C$があり、$C/C_p$とは、
$$C/C_p \coloneqq \prod_{b_i \in C, b_i \notin C_p} b_i$$
であるとする。他方、以上の相互作用の形式化は存在内における物質や情報の輸送・相互作用の記述にも同様に有用である。例えば$E$の部分集合$E_a \in E, E_b \in E$を定義して、その間の相互作用を記述すれば$E$内部での相互作用の形式化として成り立つ。

## 素朴的知能定義の選択

次に、公理的知能定義のための前段階として素朴的知能定義の選択を行う。本研究における"知能"とは、以下によって定義される概念であるとする。

"情報や物質を、外部から入力される構造、処理する構造、そして外部へ出力する構造のそれぞれを有する存在"

例えば、典型的な"ニューラルネットワーク"は外部から情報を入力される"入力層"と"中間層"と"出力層"という構造をそれぞれ有するのでこの定義に当てはまり、すなわち知能であると言える。反対に、例えば水素原子単体は電子や原子核といった構造を持つが、それらは入力構造・処理構造・出力構造のどれとも言えずこの定義に当てはまる存在ではないので、知能ではないと言える（ただ、後述するように解釈によっては水素原子単体であっても知能と見做すことができる。しかしそのような拡大解釈には有用性が乏しく、知能制作上意味がないことも同時に補足を行う）。

補足としてどのような理由をもってこの素朴的定義を選択したのかについて以下に示していく。まず、この定義は知能という概念を比較的広義に捉えたものである。例えば生物・非生物の如何を問うていないし、ただ情報や物質に対する入力・処理・出力の三つの能力を持てばそれで知能と見做すとしている。これはある高次の知能はある低次の知能の組み合わせにより構築される[e.g., 13]とした知能の構造を考慮しこのように広義に設定している。つまり例えば言語認識のような高次の知能であっても反射のような単純な機能の組み合わせによって形成されているのだから、その要素である単純な機能とは、即ち知能で

あるとした方が知能作成にあたっては有用だろうと考えるのである。

この定義方式の懸念点は、明らかに知能ではないと大多数の人間が直感するような存在であっても、解釈によっては知能であると考えることができてしまうということにある。例えばある"砂の塊"という存在を考えてみる。砂の塊は明らかに知能ではないと考えるべき存在であるとここではしておき、そして以下のように上記の知能定義を当てはめてみる。

"砂の塊は風によって入力される構造を持ち、そして内部構造を変化させ、外部へ一部の砂を出力する構造を持つ。よって、砂の塊は知能である。"

以上のように、広義に素朴的定義を行うと定義としての欠陥が生じてしまう。そしてこの定義を公理的定義に焼き直そうともこの問題は解消されないことは後のセクションの記述からも明らかである。ただ、この懸念点は実質的に無視できると考えている。なぜなら我々は知能を作成する際に、"砂の塊"のような知能作成に際して明らかに有用ではないと考えられるものは用いないからである。あるいは、そもそも知能作成においては"知能"を定義するその本質的な意義それ自体がほとんど存在しないと考えてよい。我々は明らかに"知能"を作成しようとしていて、事実作成しているはずである。本セクションのように"知能"という概念に対しては広義の枠組みだけを与え、そこから知能・非知能の境界線を議論するのではなく、実際の知能がこの枠組みからどう表現されうるのかを議論すべきであると私は考えている。以上のことは知能の定義とは有用性の追求を以てして定義する必要性自体が束縛されるのであって、有用性の乏しい"知能"を定義する意味や必要は存在しないはずであると言い換えられる。このような観点において直感的欠陥を有する広義的素朴的知能定義を公理的知能定義のために採用することは許容されうるだろう。

## 集合論的宇宙描像に基づく素朴的知能定義の公理的定義化

本セクションでは前セクションで行った素朴的知能定義を前々セクションによる集合論的宇宙描像で表現することによって公理的知能定義を行う。まず知能集合を$I \coloneqq \{v_1, v_2, \cdots, v_{l-1}, v_l\}, I \in U$として、"情報や物質を、外部から入力される構造、処理する構造、そして外部へ出力する構造のそれぞれを有する存在"という素朴的知能定義のそれぞれの語句を集合論的宇宙描像において表現する。

Def 1. "情報や物質":= $v \in U$。

Def 2. "情報や物質を外部から入力される構造":= $C_i \in C_I \subseteq v^l = v_1 \times v_2 \times \cdots \times v_{l-1} \times v_l$、ここで$C_I$は$I$の構造すべて。

Def 3. "情報や物質を処理する構造":= $C_p \in C_I$。

Def 4. "情報や物質を外部へ出力する構造":= $C_o \in C_I$。

以上の定義から"情報や物質を、外部から入力される構造、処理する構造、そして外部へ出力する構造のそれぞれを有する存在"とはすなわち、$I(C_i, C_p, C_o)$と表記される。

次に$I$が$C_i, C_p, C_o$を有する条件を記述していく。ある集合が知能集合であるために

は、これらの構造を有することが前述の素朴的知能定義から要請される。すなわち、知能集合$I$は構造$C_i, C_p, C_o$を有し、それを満たす条件を示すことが本研究における知能の定義となる。知能集合$I$と知能外集合$O := U \setminus I$はある時刻$i$からある時刻$i+1$にかけて時間写像$T_i$によって投射されることによって時間発展する。つまり、

$$T_i: I_i \cup O_i \to I_{i+1} \cup O_{i+1}, I \in U, O \in U.$$

ただし$|I_{i+1}| + |O_{i+1}| = |I_i| + |O_i|$である。ここで構造$C_i$は情報や物質を外部から入力される構造であるから、$I$が$I(C_i)$であるための条件は以下のように形式化できる。

$$|I_{i+1}| > |I_i|, |O_{i+1}| < |O_i|,$$

でありかつこの濃度変化は$O$のある部分集合 $P \in O$の受け渡しによるものである。同様に$I(C_o)$であるための条件は以下のように形式化できる。

$$|I_{i+1}| < |I_i|, |O_{i+1}| > |O_i|,$$

でありかつこの濃度変化は$I$のある部分集合 $Q \in I$の受け渡しによるものである。ただ、ある時刻間におけるある存在の濃度変化とは正負または変化なしのいずれか一つであるために、例えば入出力が同時に起こっている知能を考える場合には、この条件であると矛盾が生じる。よってより厳密に$I$が$I(C_i, C_o)$であるための条件付けは、$I$のいずれかのある部分集合$R, S$と$O$間の濃度変化で記述することであるだろう。つまり、

$$|R_{i+1}| > |R_i|, |O_{i+1}| < |O_i|, R \in I,$$

かつ同時に、

$$|S_{i+1}| < |S_i|, |O_{i+1}| > |O_i|, S \in I,$$

であれば$I$は$I(C_i, C_o)$であるとする。ただし、$I, O$の同じ要素が同時刻間に入力されかつ出力されることはないとする。このように出入力構造を定義すれば上述のような矛盾は生じない。例えば$R, S$の入出力される要素がそれぞれ1つのみであればそれはある時刻間には入力もしくは出力されるのみである。そして$I$におけるいずれかの部分集合$R, S$と$O$間で条件に適する濃度変化があれば、$I$は$I(C_i, C_o)$であると見做せるとしているために、$I, O$におけるある要素が入出力されていればその$I$は$C_i, C_o$を有するのである。最後、$I$が$I(C_p)$であるための条件は、$I$のある部分集合$T, V$が以下のように時間発展することである。

$$|T_{i+1}| > |T_i|, |V_{i+1}| < |V_i|, T \in I, V \in I,$$

もしくは

$$|T_{i+1}| < |T_i|, |V_{i+1}| > |V_i|.$$

ここで$I$の部分集合は$T$と$V$以外にも存在しうるために、上述のような集合間の保存則は成り立たない。これらの変化は外部集合$O$との要素の移動によるものではなく、知能集合内部での要素の再構成によるものである。すなわち時刻間での変化は$I$の内部構造に起因するものである。このような定義により、処理構造$C_p$は情報・物質の分配、変換、記憶、再構築などの振る舞いを包括的に含んだ知能の内部的活動に対応する形式化として成り立つ。

## 公理的知能定義によるある知能の形式化と再解釈

　　　　本セクションではこれまでに行ってきた公理的知能定義を実在する具体的な知能に当てはめ、その機能や作用について再解釈を与える。ヘブ則型非最適化 NNs による MNIST 判定機構[13, 21]、逆伝搬型最適化 NNs による MNIST 判定機構 [e.g., 19, 21]、そしてアメフラシ鰓開閉機構 [e.g., 20]の各知能集合について記述を行う。このことによりこれらの知能の差異や生物模倣性の程度、そして生物模倣 AI を作成する上での禁足事項について議論する。まず、それぞれの集合としての区分や要素、そして構造について表 1 にまとめ、そして時間発展に伴ってそれらがどのように入力・処理・出力が行われるかについて表 2 にまとめる。表 2 では学習過程とテスト・実動過程に動作過程を分けて記述し、入力・処理・出力はその順に別々の時刻間に起こるとしている。

|  | ヘブ則型非最適化 NNs による MNIST 判定機構[13] | 逆伝搬型最適化 NNs による MNIST 判定機構 [e.g., 19] | アメフラシ鰓開閉機構 [e.g., 20] |
|---|---|---|---|
| 集合区分 | NNs や他機能を定義するプログラムコードやそれによって動作するコンピュータ。10 並列 NNs。 | NNs や他機能を定義するプログラムコードやそれによって動作するコンピュータ。非並列 NNs。 | アメフラシの生体内で鰓の開閉に関与する細胞群。非並列神経系。 |
| 要素 | コンピュータの各演算機器。仮想的には、プログラムコードによって定義される各ノードやその結合、ヘブ則による重み更新構造、ベクトルノルム計算構造。 | コンピュータの各演算機器。仮想的には、プログラムコードによって定義される各ノードやその結合、加えて損失関数や逆伝搬による重み更新機能、ソフトマックス関数。 | アメフラシ体内の単一細胞。それによって形作られる神経細胞など。具体的には神経系、感覚受容細胞群、運動神経細胞群、筋組織、関連する神経伝達物質を含む。 |
| 構造$C_i$ | 入力層。 | 入力層。損失関数。出力層。 | 体表面の感覚受容器（例えば触覚や水流感知）を介した外部刺激の受容構造。 |
| 構造$C_p$ | 中間層。 | 中間層。逆伝搬機能。 | 体内神経系やそれを維持する各細胞による構造。神経伝達物質産生・放出機能。ヘブ則によるシナプス強化機能。 |
| 構造$C_o$ | 出力層。ベクトルノルム計算による正誤判定機能。 | 出力層。ソフトマックス関数。ラベル数に応じたノード数の調整有。 | 鰓の開閉機能。具体的には筋細胞の収縮と弛緩により鰓が開閉する。閾値構造有。 |

表 1. 各知能集合の集合区分、要素、各構造のまとめ。

| 学習過程時間変化 | ヘブ則型非最適化 NNs による MNIST 判定機構[13] | 逆伝搬型最適化 NNs による MNIST 判定機構 [e.g., 19] | アメフラシ鰓開閉機構 [e.g., 20] |
|---|---|---|---|
| $T_i: I_i \xrightarrow{C_i} I_{i+1}$ | 並列 NNs へのラベル別での学習データの入力層への入力。中間層への応答。 | 学習データの入力層への入力。中間層への応答。 | 外界からの物理的刺激が感覚受容細胞群により検知。神経細胞群への応答。 |
| $T_i: I_{i+1} \xrightarrow{C_p} I_{i+2}$ | 中間層の順伝搬。ヘブ学習による重み更新。 | 中間層の順伝搬。 | 神経細胞群間での信号の順伝搬。ヘブ則によるシナプス結合の強化。短期学習や長期学習、馴化含む。 |
| $T_i: I_{i+2} \xrightarrow{C_o} I_{i+3}$ | 出力層の応答（Itoh, 2024b では意味のある応答なし）。 | 出力層の損失関数への応答。 | 未学習の判断基準（閾値）による鰓の開閉応答。 |
| $T_i: I_{i+3} \xrightarrow{C_i} I_{i+4}$ | - | 損失関数による出力層への応答。 | - |
| $T_i: I_{i+4} \xrightarrow{C_p} I_{i+5}$ | - | 中間層の逆伝搬と重み更新。 | - |
| テスト・実動過程時間変化 | ヘブ則型非最適化 NNs による MNIST 判定機構[13] | 逆伝搬型最適化 NNs による MNIST 判定機構 [e.g., 19] | アメフラシ鰓開閉機構 [e.g., 20] |
| $T_i: I_i \xrightarrow{C_i} I_{i+1}$ | テストデータの入力層への入力。中間層への応答。 | テストデータの入力層への入力。中間層への応答。 | 外界からの物理的刺激が感覚受容細胞群により検知。神経細胞群への応答。 |
| $T_i: I_{i+1} \xrightarrow{C_p} I_{i+2}$ | 中間層の順伝搬。出力層への応答。 | 中間層の順伝搬。出力層への応答。 | 神経細胞群間での信号の順伝搬。 |
| $T_i: I_{i+2} \xrightarrow{C_o} I_{i+3}$ | 出力層の応答。ベクトルノルム計算による正誤判定。 | 出力層の応答による正誤判定。 | 学習された判断基準（閾値）による鰓の開閉応答。 |

表２．各知能集合の時間発展に伴う集合内変化のまとめ。入力・処理・出力はその順に別々の時刻間に起こるとする。$\xrightarrow{C}$記号はある構造$C$がその時間発展において集合の変化に関与することを示す。

表 1.2 で示した通り、ヘブ則型非最適化 NNs は局所的なヘブ則による重み更新構造とベクトルノルム計算構造を学習構造・正誤判定構造として持つ。このため、その学習過程は局所的な重み更新を行うのみで、損失関数は存在しない。一方、逆伝搬型最適化 NNs は非局所的な損失関数および逆伝搬機能に基づいて重みを更新し、一般に出力構造としてソフトマックス関数を採用している。これらに対し、生物的知能であるアメフラシの鰓開閉機構は、神経細胞群とその相互作用、筋細胞群の運動を通じた鰓の開閉を出力としており、閾値を含む内部基準に基づいて活動する。表 2 の時間発展過程の分析からは、ヘブ則型非最適化 NNs とアメフラシの神経回路が類似した特徴を示すことが明らかになる。どちらも外部からの入力に応じて局所的かつ自己組織的な NNs や神経系の構造変化が起こり、非局所的な誤差関数を必要とせず学習を行うことができる。逆伝搬型最適化 NNs は外部的な損失関数を必須としておりそのために学習過程において必要な時間ステップが他二つよりも多い。これらの点で生物的模倣性から大きく乖離していることが示されている。

しかし、ヘブ則型非最適化 NNs にも非生物的な機構が存在する。Itoh (2024b) [13] のモデルでは、並列 NNs を用いた構成が採用されており、それぞれの NNs が特定のラベルに対して個別に訓練される。こうした並列構造は、ヘブ則の有用性を簡便に示すための工学的・実利的利点を持つ一方で、生物の神経系における非並列的かつ統合的な情報処理の様式とは乖離がある。生物神経系では複数の機能が単一の神経構造内で処理されるのが一般的であり、並列的にラベルごとに構造を分けるという設計は明らかに人工的・非生物的である。さらに、ヘブ則という学習規則そのものが生物的に妥当であったとしても、それが用いられる目的——すなわち文字認識という高次機能——自体が、そもそも生物の持つ反射的行動機構とは本質的に異なるものである。アメフラシの鰓開閉のような生物的反応は、特定の刺激に対する特定の身体運動を誘発する低次の反射系であり、そこに"分類"や"認識"といった抽象的情報処理は介在しない。したがって、単純な構造による設計では文字認識を目的に設定していると、たとえ生物模倣的であると称していても、本質的には非生物的な処理目的を前提としていることになる（つまり、生物模倣による文字認識の実装は真に生物が文字認識を行う方式を採用しなければならない）。このように、ヘブ則型非最適化 NNs はその局所性や非最適化構造において生物的特徴を一定備えてはいるものの、その並列構造や機能目的において明確に非生物的な特徴を含んでいる。ゆえに、生物模倣性を高め最終的に生物模倣 AI を作成するためには、単にヘブ学習則を導入することに留まらず、その適用方法や構造設計、さらには目的に至るまで再検討する必要があるだろう。以上の分析を踏まえると、生物模倣型 AI 設計においては非局所的な損失関数や逆伝搬のような機構は禁則とすべきであり、局所的で自己組織的なヘブ則的学習構造を積極的に採用することが望ましい。ただし、その実装においては NNs 並列構造といったその他の非生物的要素に対しても同様に慎重な設計判断が求められ、目的によっては禁則とすべきであるだろう。また、人間模倣型 AI や AGI の開発はこのような知能の公理的理解を人間や AGI に対して進めることにより進展することは明らかであり、本研究はその基礎的理論を形成する礎となりうるだろう。

追加の議論としてより高度な公理的知能定義やその記述について考える。本研究では一貫して素朴的な知能の定義・理解をより公理的に形式化が可能な定義に置き換え理解する試みを行ってきた。しかし本研究では確かに知能という名のフレームワークに対しては形式的記述を行ったが、ある知能の具体的な内容・実動については自然言語によって記述するに形式化の程度は留まっている。今後はある知能の、そしてどのような具体的な知能の内容・実動をも完全な公理的形式化によって記述することを目指すべきだろう。このことが達成されれば、より公理的な知能の理解だけではなく、知能の回路的記述やそれに伴う知能のハードウェア的実装をも可能とするかもしれない。

**考察**
**公理的知能定義の任意性と圏論への拡張によるその例示**
　　　　本研究では集合論的形式化による知能の公理的定義を進めたが、この公理的定義の方法論については依然として大きな任意性を持つことを確認する。つまり、本研究による定義やその方法論は一例であって、より優れた理論を開発しても良いし、全く異なる数学理論や素朴的知能定義選択によって記述しても良い。それらの必要性は、研究目的セクションにおいて記述した通り、その定義の有用性によって要請されるだろう。また本考察では、集合論ではなく圏論によっても知能定義が行うことが可能だということをその任意性の例示として簡単に記述する。集合論的宇宙観やそれによる存在・構造の定義や素朴的知能定義選択は前提として受け入れたうえで、知能圏$\mathcal{I}$と時間圏$\mathcal{T}$の二つの圏の射や関手から知能やその時間発展についてより発展的なより高階の構造的定式化を行う。知能とは入力・処理・出力の三構造を有する対象として、圏$\mathcal{I}$により次のように定式化される：

$$\mathcal{I} = \text{Cat}_{\text{Intelligence}}, \text{Obj}(\mathcal{I}) = \{I = (C_i, C_p, C_o)\}$$

$$\text{Mor}_\mathcal{I}(I, J) = \{f: I \to J | f = (f_i, f_p, f_o), f_k: C_k^I \to C_k^J, k \in \{i, p, o\}\}$$

ここで、対象$I$はある知能を意味し、その構造は入力集合$C_i$、処理集合$C_p$、出力集合$C_o$によって与えられる。また、$I$はすべての構造的知能対象$(C_i, C_p, C_o)$を代表する。$f: I \to J$は、ある知能から別の知能への構造写像（例えば学習、変換、模倣など）であり、構造ごとに個別の写像$f_k$を持つ構造保存射である。一方、時間の進行は別の圏として定義される。時刻を対象とし、時刻の発展を射とする時間圏$\mathcal{T}$を次のように定める：

$$\mathcal{T} = \text{Cat}_{\text{Time}}, \text{Obj}(\mathcal{T}) = \{t_i\}$$

$$\text{Mor}_\mathcal{T}(t_i, t_j) = \begin{cases} \{\tau_{ij}\} & \text{if } i \leq j \\ \emptyset & \text{otherwise} \end{cases}$$

ここでの射$\tau_{ij}: t_i \to t_j$は、時刻$t_i$から$t_j$への時間的経過を表し、合成$\tau_{ij} \circ \tau_{jk} = \tau_{ik}$によって時間の連続性を満たす。このとき、$I$に関する時間圏$\mathcal{T}$から知能圏$\mathcal{I}$への時間関手$F_I: \mathcal{T} \to \mathcal{I}$によって、知能の時間発展が定式化される。具体的には：

- 各時刻$t_i \in \text{Obj}(\mathcal{T})$に対して、各時刻での知能$I_i = F_I(t_i) \in \text{Obj}(\mathcal{I})$を対応させ、
- 各射$\tau_{ij}: t_i \to t_j \in \text{Mor}(\mathcal{T})$に対して、時間発展に伴う知能の構造変化$f_{ij} = F_I(\tau_{ij}): I_i \to$

$I_j \in \text{Mor}(\mathcal{J})$ を対応させる。

これにより、関手 $F_I$ は以下の条件を満たす。

恒等射の保存：$F_I(\text{id}_{t_i}) = \text{id}_{I_i}$,

合成の保存：$F_I(\tau_{jk} \circ \tau_{ij}) = F_I(\tau_{jk}) \circ F_I(\tau_{ij})$.

なお、知能対象が異なれば、それぞれに対応する別の時間関手 $F: \mathcal{T} \to \mathcal{J}$ を定義することができる。このようにして、各知能の時間発展は対応する時間関手 $F$ によって一貫して記述される。また、射 $f_{ij} = F(\tau_{ij})$ が知能圏内の射であることから、各時間ステップにおける知能の状態変化は、より一般の変換（学習など）を射として定義可能であるだろう。さらに、複数の知能体系（たとえば生物知能と人工知能）をそれぞれ別の知能圏 $\mathcal{J}_\text{Bio}, \mathcal{J}_\text{AI}$ として定義し、それらの間の関係を知能間関手 $G: \mathcal{J}_\text{Bio} \to \mathcal{J}_\text{AI}$ によって記述すれば、生物的知能構造の人工的模倣過程も数学的に定式化されうる。これは例えば表 1 は知能圏 $\mathcal{J}$ とその射、そして関手 $G$ によって、表 2 は知能圏 $\mathcal{J}$ と時間圏 $\mathcal{T}$ とその射、そして関手 $F, G$ もしくは合成関手 $H = F \circ G$ によってそれぞれ記述可能であることを示唆し、上述の完全な公理的知能形式化のための手段になりうる。加えて、知能圏における射や知能圏間関手は学習など時間発展に起因する変化だけでなく、模倣・翻訳・抽象化といった非時間的な機能変換も含んでいる。これらを合成することで生物知能から完全生物模倣 AI への構造的写像など、大規模かつ高高度な構造変換も射・関手として一括して記述可能である可能性を秘める。また、集合論的枠組みでは具体的な要素の移動に着眼点が置かれたのに対し、圏論的枠組みではそれぞれの知能の機能がどのように対応付けられ変換されうるかに対して着眼されるはずである。そのように多様な価値観において公理的知能定義が行われることは、知能のより高度な理解に繋がり、ひいては人間模倣 AI や AGI の開発に大きな貢献をするだろう。

### 知能の活動度について

本研究における公理的知能定義は、知能とは時間発展を伴う集合間相互作用を要件とするものである。したがって、時間発展の過程において他集合との相互作用を一切伴わない集合は、たとえ構造として入力・処理・出力機能を備えていたとしても、その時刻においては知能集合とは見做されない。すなわち、ある集合が知能構造を有していても、何らかの要因によって時間的にその構造が実際に機能していない場合、その集合は当該時間帯においては知能として振る舞っていないということになる。例えば人工知能において、内部に知能的構造を備えていても、コンピュータが通電されておらず、外部との意味ある相互作用が生じていないのであれば、それは知能として発現していない状態であり、本定義においては知能たりえない。そうすると本研究の知能定義は、知能存在において機能するための物質的・環境的前提条件が満たされていることを暗に仮定しているとも解釈できる。そしてこの点をより厳密に取り扱おうとするならば、知能構造の定義に加えて、それが知能として実際に機能するための条件群もまた、定義の中に明示的に組み込まれるべきであろう。この観点は、知能の"活動度（activity）"とでも呼ぶべき性質を導入することで、より鮮明に理解可能

となる。ここで言う活動度とは、ある知能構造が任意の時刻間において、どの程度知能として実際に機能していたか、すなわち他集合との相互作用を実行していたかを示す量的指標である。例えば、あるプログラムによって規定された知能は、その構造を保持していても、起動されていなければ知能的機能を果たさない。したがって、起動していない時間帯における人工知能存在は、知能としての活動度がゼロであると評価されうる。一方、生物はその誕生から死に至る時間の中で、一貫して外部との相互作用を通じた知能的活動を行っており、その活動度は高いと考えられる。ただ例えば睡眠時と覚醒時ではそれらの活動度には差があるかもしれない。そしてこのようにこの活動度という概念は、生物知能と非生物知能、人間知能と非人間知能、さらにはAGIと非AGIといった知能の種別を定義・区分する際において、有用な補助的概念となる可能性が高いと考えられる。

### 知能以外の概念の公理的定義

本研究において行った公理的知能定義の手順は、知能以外の概念に対しても適用可能なはずである。つまり例えば高度知能開発には不可欠であると考えられかつ知能と同様に本来素朴的である"意識"や"感情"といった概念を"集合論的宇宙描像の定義→素朴的概念選択→集合論的宇宙描像に基づく素朴的概念の公理的定義化"という手順を踏むことにより公理的に再解釈するのである。これは知能における部分的な機能として位置付けられるかもしれないし、もしくは並列的に取り扱われる個別の概念であるかもしれない。そしてこれは本研究において論じてきたように、方法論を含めて複数の定義を保持可能であるだろう。集合論で記述しても良いし、前々セクションのように圏論で記述しても良いし、他の理論（例示するなら、トポス理論など）を用いるのでももちろん良い。そして"意識"や"感情"以外にも必要であり有用であるのならばどのような概念であっても公理的に定義すべきだろう。あるいは、本研究はある意味で今まで人々の間で形而上論的に扱われてきた一部の概念を形而下論として扱うための手法を提供するものであるとも言えるだろう。つまりこれは、カントが"物自体[22]"と定義付け、そしてヴィトゲンシュタインが"語りえぬもの[23]"として沈黙を促した領域の一部を、"語りえる現象"とする試みであるかもしれない。

### 引用